\documentclass{article}
\usepackage{amsmath}
\usepackage{amssymb}
\usepackage[cmtip,arrow]{xy}
\usepackage{pb-diagram,pb-xy}

\def\resize#1\tosizeof#2{{\setbox0=\hbox{$\displaystyle{#2}$}\hbox to\wd0{\hss$\displaystyle{#1}$\hss}}}
\title{The Fast Tree Transform}
\author{Paul Burchard}
\date{June 1, 2013}
\begin{document}
\maketitle

\begin{abstract}
The {\em Fast Tree Transform\/} is a locally-aggregated transform defined for arbitrary graph geometries that generalizes both the Hadamard frequency transform and the Haar wavelet transform on the circle.
The pure frequency part of the transform approximately diagonalizes the graph Laplacian; this fact makes it useful as a preconditioner in geometric optimization problems.
The full transform provides an only slightly-redundant representation of signals on the graph that allows sparse representation of signals that include both edges and textures.
\end{abstract}

\section{Concepts}

\begin{figure}[h]
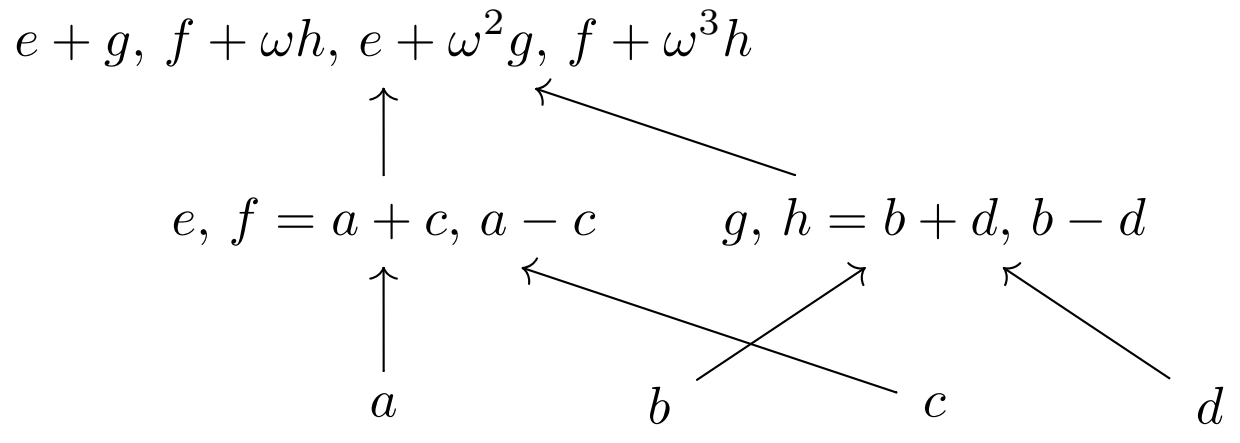

\begin{center}
$\displaystyle
\begin{diagram}
     \node{\resize{e + g,\, f + \omega h,\, e + \omega^2 g,\, f + \omega^3 h}\tosizeof{XXXXXXXXX}} \\
     \node{\resize{e,\,f = a + c,\, a - c}\tosizeof{XXXXXXX}} \arrow{n} \node[2]{\resize{g,\,h = b + d,\, b - d}\tosizeof{XXXXXXX}} \arrow{wnw} \\
	\node{a} \arrow{n} \node{b} \arrow{ne} \node{c} \arrow{wnw} \node{d} \arrow{nw}
\end{diagram}
$
\end{center}
\caption{Fast Fourier Transform ($\omega$ is a fourth root of one).}
\label{fig.fft}
\end{figure}

\begin{figure}[h]
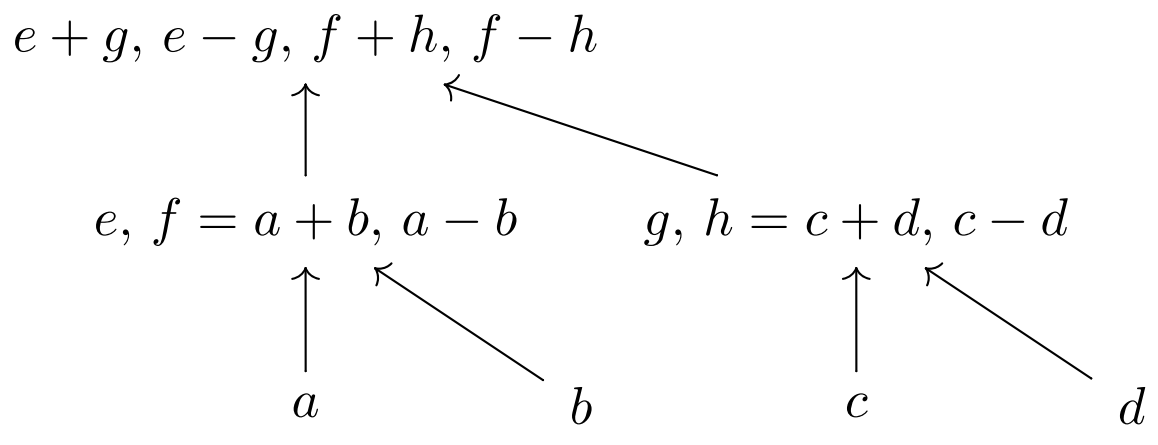

\begin{center}
$\displaystyle
\begin{diagram}
     \node{\resize{e + g,\, e - g,\, f + h,\, f - h}\tosizeof{XXXXXXXXX}} \\
     \node{\resize{e,\,f = a + b,\, a - b}\tosizeof{XXXXXXX}} \arrow{n} \node[2]{\resize{g,\,h = c + d,\, c - d}\tosizeof{XXXXXXX}} \arrow{wnw} \\
	\node{a} \arrow{n} \node{b} \arrow{nw} \node{c} \arrow{n} \node{d} \arrow{nw}
\end{diagram}
$
\end{center}
\caption{Fast Tree Transform (simplest case).}
\label{fig.ftt}
\end{figure}

What's wrong with the Fast Fourier Transform?  The first problem is that its efficiency depends on global structure, which is not available for arbitrary geometries.
As shown in Figure~\ref{fig.fft}, at each level the algorithm combines values across the largest strides of the geometry.
To get a frequency transform based on local structure, that could be generalized to arbitrary geometries, we have to be able to combine values locally and aggregate up.
Fortunately, as shown in Figure~\ref{fig.ftt}, in the same example geometry, there is a simple local aggregation of sums and differences of values that yields an orthogonal basis displaying the full range of frequencies, which we will call the {\em Fast Tree Transform}.

In the example of Figure~\ref{fig.ftt}, the basis is:
\[
  \begin{matrix}
    + & + & + & + \\
    + & + & - & - \\
    + & - & + & - \\
    + & - & - & +
  \end{matrix}
\]
As can be seen from the basis, in this simple case of a circular geometry of dyadic size, this locally aggregated transform reduces to the {\em Hadamard transform}, using the basis ordering first defined by Paley~\cite{bib.fino.hadamard}.  The choice of ordering is not just cosmetic, as it is key to keeping the aggregation local for the generalization to arbitrary geometries.

To see why this should be considered a frequency transform, although the basis functions of this transform are not guaranteed to be eigenfunctions of the graph Laplacian $\Delta$, we can compute the spectral value of such a function $\psi$ as
\[
  \frac{\psi \cdot \Delta \psi}{\psi \cdot \psi} = - \frac{\sum_e (\nabla \psi(e))^2}{\sum_v \psi(v)^2} ,
\]
where the sums in the second form are over edges and vertices, respectively.
This is just the ratio of the harmonic energy to the $L^2$-norm.
In the example of Figure~\ref{fig.ftt}, we find the spectrum of the basis vectors to be (respectively):
\[
  \begin{matrix}
    0 \\
    -2 \\
    -4 \\
    -2
  \end{matrix}
\]
In this small example, we happen to reproduce the spectrum of the Laplacian exactly.
A bit more generally, for circular geometries of larger dyadic size, it is not hard to see that the spectral values of the basis functions fill out the spectrum of the Laplacian linearly, while the true spectrum fills itself out quadratically.

\begin{figure}[h]
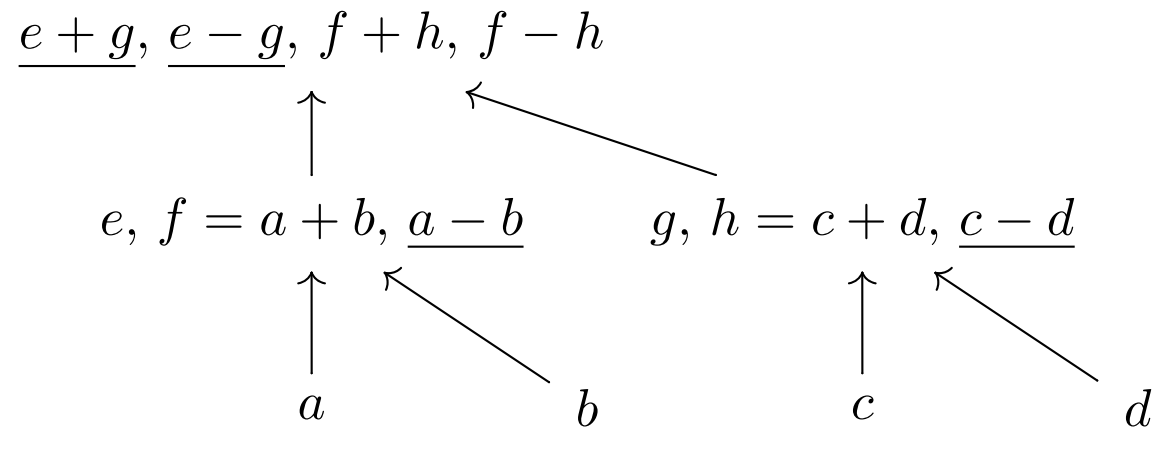

\begin{center}
$\displaystyle
\begin{diagram}
     \node{\resize{\underline{e + g},\, \underline{e - g},\, f + h,\, f - h}\tosizeof{XXXXXXXXX}} \\
     \node{\resize{e,\,f = a + b,\, \underline{a - b}}\tosizeof{XXXXXXX}} \arrow{n} \node[2]{\resize{g,\,h = c + d,\, \underline{c - d}}\tosizeof{XXXXXXX}} \arrow{wnw} \\
	\node{a} \arrow{n} \node{b} \arrow{nw} \node{c} \arrow{n} \node{d} \arrow{nw}
\end{diagram}
$
\end{center}
\caption{Haar Wavelet Transform coefficients (underlined) as subset of Fast Tree Transform computation.}
\label{fig.wavelet}
\end{figure}

The second problem with the Fast Fourier Transform is that it does not allow for efficient representation of typical spatial signals, which contain both spatially-localized discontinuities (edges in images, onset of a tone in sounds) and frequency-localized patterns (textures in images, pitch and timbre in sounds).
The traditional answer to this problem has been to use {\em fast wavelet transforms\/} which combine spatial and frequency localization,
but these are only efficient at representing discontinuities and not patterns.
The reason both of these approaches fail is that sparse representation of such spatial signals requires some redundancy.
If we keep intermediate results and not just the final output, however, the Fast Tree Transform carries exactly the minimal redundancy to capture both types of features.
Indeed, with a nearly linear number of coefficients, it contains both the Hadamard frequency transform as the root-level coefficients,
and the wavelet transform in the Haar basis as a different subset of the coefficients across levels of the aggregation tree (see Figure~\ref{fig.wavelet}).
The remaining coefficients correspond to patterns that have some spatial localization.

A sparse representation of a spatial signal can thus be obtained by appropriate thresholding of the coefficients of the Fast Tree Transform, including intermediate coefficients.
To reconstruct the signal from such a representation, we begin computing the inverse transform by local aggregation, starting from the pure frequency coefficients, but at each stage overriding any intermediate coefficient with the previously computed thresholded coefficient if it is non-zero.

\section{Construction}

Now that we have the concept of a frequency transform based on local aggregation, we are ready to generalize it to arbitrary geometries.
Given an arbitrary graph, we choose a {\em low-stretch spanning tree\/} for it, which can be discovered in nearly linear time~\cite{bib.alon.lowstretch,bib.abraham.lowstretch}.
We then recursively break this spanning tree into two nearly equal parts along an edge,
forming a binary classification of the vertices.
This classification is represented by an ordered binary tree, called the {\em localizing tree\/} for the graph,
whose leaves correspond to vertices of the graph, and whose branches correspond to edges of the spanning tree.
The transform will then accumulate sums and differences of values up this localizing tree.

In order to define the Fast Tree Transform, an ordered binary tree $\alpha$ must be decorated with some additional information.
This will involve the notion of a {\em dual tree\/} $\alpha^*$, which will end up being the ordered binary tree for the inverse transform.
The required decoration at each branch is an {\em intersection\/} of the dual trees of its left and right branches, by which we mean
an ordered binary tree together with an ordered embedding of the left and right branch dual trees into it, such that the every leaf is in at least one of the two embedded trees (an embedding must map leaves to leaves and branches to branches, preserving order and ancestry relations, in a way that results in a topological embedding of trees).

\begin{figure}[h]
\begin{center}
\setlength{\unitlength}{0.0005\textwidth}
\begin{picture}(1700,500)(0,-300)
  \thinlines
  \put(200,100){\makebox(0,0)[c]{$\alpha$}}
  \put(200,0){\line(-1,-1){100}}
  \put(200,0){\line(1,-1){100}}
  \put(100,-100){\line(-1,-1){100}}
  \put(100,-100){\line(1,-1){100}}

  \put(500,100){\makebox(0,0)[c]{$\beta$}}
  \put(500,0){\line(-1,-1){100}}
  \put(500,0){\line(1,-1){100}}

  \put(900,100){\makebox(0,0)[c]{$\alpha \cap \beta$}}
  \put(900,0){\line(-1,-1){100}}
  \put(900,0){\line(1,-1){100}}
  \put(800,-100){\line(-1,-1){100}}
  \put(800,-100){\line(1,-1){100}}
  \put(900,20){\line(-1,-1){220}}
  \put(900,20){\line(1,-1){120}}

  \put(1450,100){\makebox(0,0)[c]{$\alpha \wedge \beta$}}
  \put(1450,0){\line(-3,-2){150}}
  \put(1450,0){\line(3,-2){150}}
  \put(1300,-100){\line(-1,-1){100}}
  \put(1300,-100){\line(1,-1){100}}
  \multiput(1200,-200)(-8,-8){13}{\circle*{2}}
  \multiput(1200,-200)(8,-8){13}{\circle*{2}}
  \multiput(1600,-100)(-8,-8){13}{\circle*{2}}
  \multiput(1600,-100)(8,-8){13}{\circle*{2}}
\end{picture}
\end{center}
\caption{Tree intersection and wedge product.}
\label{fig.intersect}
\end{figure}
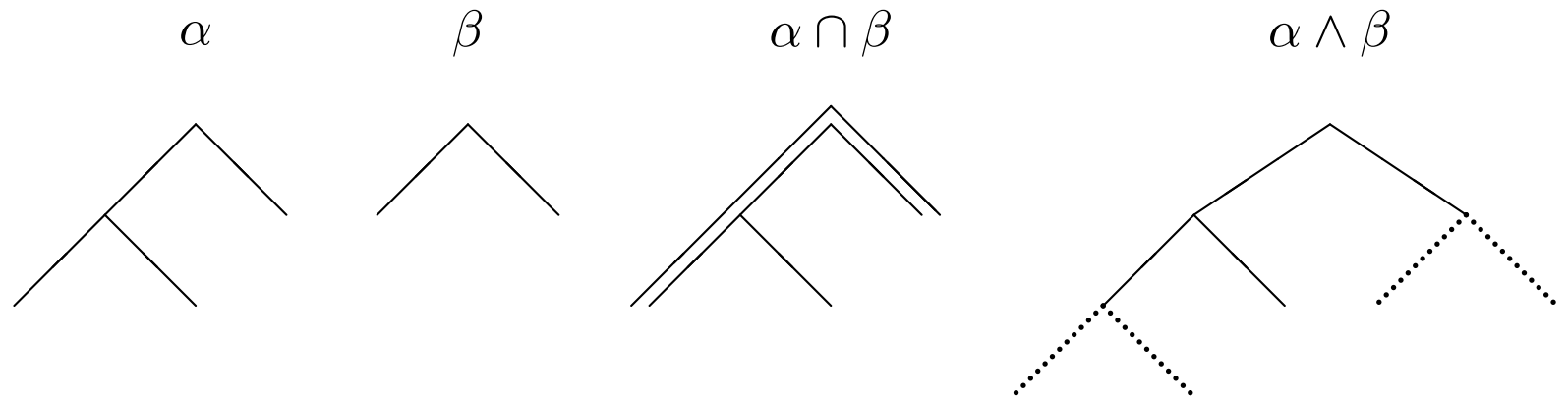

To define the dual, we will need some notation.
Given two ordered binary trees $\alpha$ and $\beta$, let $\alpha \vee \beta$ denote the (unique) tree with a new root whose left branch is $\alpha$ and right branch is $\beta$,
and let $\alpha \wedge \beta$ denote a (not necessarily unique) intersection of the two trees with each duplicate leaf split into a branch with two leaves (see Figure~\ref{fig.intersect}).
Both operations are additive in the number of leaves.
Recursively, then, the dual of a leaf is just the leaf itself,
and the dual of a branch $(\alpha \vee \beta)^* = \alpha^* \wedge \beta^*$ is the wedge product of the duals determined by the decoration of that branch.
We will show below that the dual of the dual is indeed the original tree.
In the example of Figure~\ref{fig.ftt} of a fully balanced binary tree, all the branches are self-dual, and the intersection of duals at each branch is just the equality of the left and right dual trees.
For an example of intersecting non-equal trees, see Figure~\ref{fig.intersect}.

Given a decorated ordered binary tree $\alpha$, we now define the {\em Fast Tree Transform} $\widehat{\ }^\alpha$, an orthogonal linear transform which can be computed in nearly linear time, on vectors of values corresponding to the ordered leaves of the tree $\alpha$.
Moreover, the values of the transform correspond to the leaves of the dual tree $\alpha^*$, and the Fast Tree Transform $\widehat{\ }^{\alpha^*}$ corresponding to the dual tree provides the inverse transform to $\widehat{\ }^\alpha$.
The transform is defined recursively by local aggregation.
For a tree with one leaf, the transform is the identity.
For a tree $\alpha \vee \beta$ with dual $(\alpha \vee \beta)^* = \alpha^* \wedge \beta^*$ corresponding to a chosen intersection of $\alpha^*$ and $\beta^*$,
we compute the transform of the concatenation $x \vee y$ of vectors corresponding to leaves of $\alpha$ and $\beta$ as follows:
Line up the values of $\hat{x}^\alpha$ and $\hat{y}^\beta$ with the leaves of $\alpha^*$ and $\beta^*$ in the chosen intersection,
and replace leaves which have values $\hat{x}^\alpha_i$ and $\hat{y}^\beta_j$ from both dual trees with an ordered pair of values $\frac{1}{\sqrt{2}}\left(\hat{x}^\alpha_i + \hat{y}^\beta_j\right)$ and $\frac{1}{\sqrt{2}}\left(\hat{x}^\alpha_i - \hat{y}^\beta_j\right)$.
The resulting values thus correspond to leaves of $\alpha^* \wedge \beta^* = (\alpha \vee \beta)^*$.
Moreover, the coefficients of this set of values clearly form an orthonormal basis. 

The choice of decoration is important for efficiency; in particular, the chosen intersections should allow as much overlap as possible.
In the opposite extreme, where we choose at every branch $\alpha^* \wedge \beta^* = \alpha^* \vee \beta^*$ to have no overlap at all, the resulting transform is just the identity, which obviously offers no discrimination of the spectrum of the Laplacian.
This is why we need a well-localized tree for the geometry, which is reasonably well-balanced at all levels, allowing a lot of overlap.

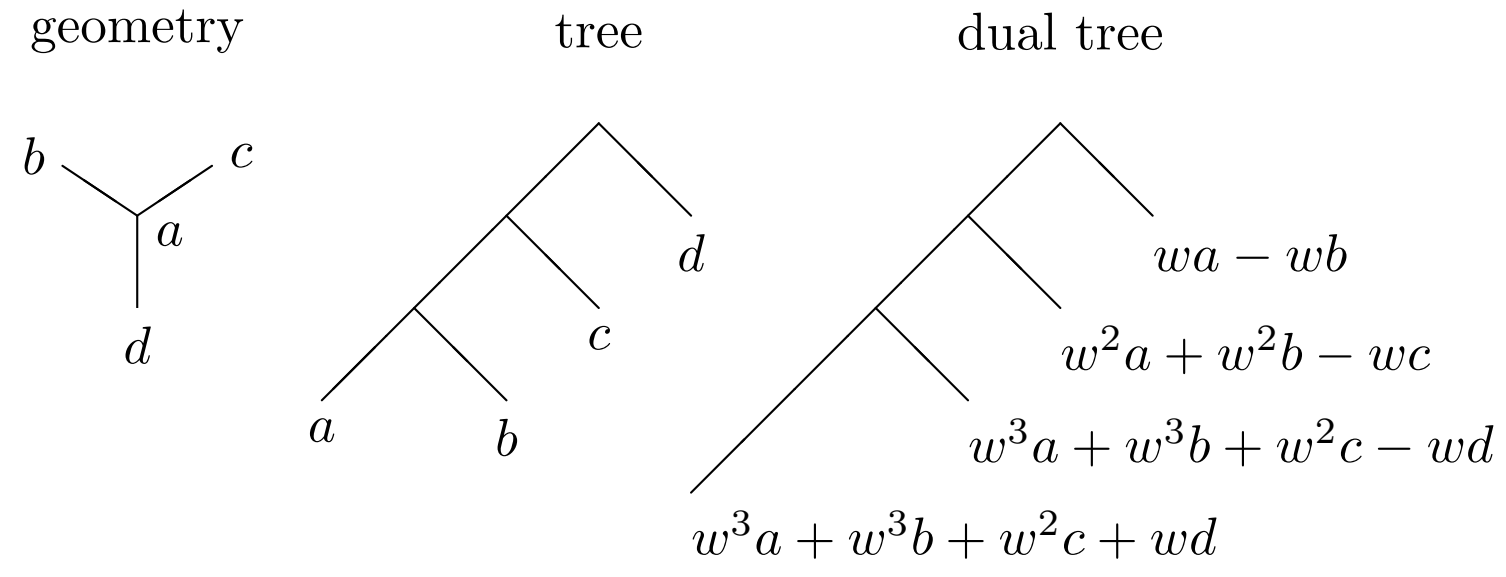
\begin{figure}[h]
\begin{center}
\setlength{\unitlength}{0.0005\textwidth}
\begin{picture}(1400,550)(0,-450)
 \thinlines
 \put(100,100){\makebox(0,0)[c]{geometry}}
 \put(120,-120){\makebox(0,0)[l]{$a$}}
 \put(100,-100){\line(0,-1){100}}
 \put(100,-220){\makebox(0,0)[t]{$d$}}
 \put(100,-100){\line(-3,2){81}}
 \put(0,-35){\makebox(0,0)[r]{$b$}}
 \put(100,-100){\line(3,2){81}}
 \put(200,-35){\makebox(0,0)[l]{$c$}}

 \put(600,100){\makebox(0,0)[c]{tree}}
 \put(600,0){\line(-1,-1){300}}
 \put(300,-320){\makebox(0,0)[t]{$a$}}
 \put(600,0){\line(1,-1){100}}
 \put(700,-120){\makebox(0,0)[t]{$d$}}
 \put(500,-100){\line(1,-1){100}}
 \put(600,-220){\makebox(0,0)[t]{$c$}}
 \put(400,-200){\line(1,-1){100}}
 \put(500,-320){\makebox(0,0)[t]{$b$}}

 \put(1100,100){\makebox(0,0)[c]{dual tree}}
 \put(1100,0){\line(-1,-1){400}}
 \put(700,-420){\makebox(0,0)[tl]{$w^3 a + w^3 b + w^2 c + w d$}}
 \put(1100,0){\line(1,-1){100}}
 \put(1200,-120){\makebox(0,0)[tl]{$w a - w b$}}
 \put(1000,-100){\line(1,-1){100}}
 \put(1100,-220){\makebox(0,0)[tl]{$w^2 a + w^2 b - w c$}}
 \put(900,-200){\line(1,-1){100}}
 \put(1000,-320){\makebox(0,0)[tl]{$w^3 a + w^3 b + w^2 c - w d$}}
\end{picture}
\end{center}
\caption{Fast Tree Transform for star geometry (where $w = \frac1{\sqrt{2}}$).}
\label{fig.star}
\end{figure}

Nevertheless, if the geometry itself forces a certain amount of imbalance, that is OK.
In particular, suppose the geometry is a star, that is, a single branch connected to a number of leaves.
This geometry is its own spanning tree.
Repeatedly splitting it into as close as half as possible results in a localizing tree which is actually a maximally imbalanced binary tree (see Figure~\ref{fig.star}).
This tree is self-dual.
In the case of a star geometry with~10 leaves, for example, following the pattern of Figure~\ref{fig.star},
the spectral values of the resulting basis functions are approximately
\[
  [-0.90,\, -0.93,\, -0.93,\, -0.94,\, -0.95,\, -0.97,\, -1.08,\, -1.36,\, -2.02,\, -3.46,\, -6.50],
\]
while the exact spectrum of the graph Laplacian is
\[
  [0,\,-1,\,-1,\,-1,\,-1,\,-1,\,-1,\,-1,\,-1,\,-1,\,-11],
\]
a good approximation.
The degeneracy of the exact spectrum, with repeated eigenvalues of $-1$, reflects the fact that all the edges in the star geometry are equally weighted.

Finally, as promised, we must show how to decorate the dual tree, in such a way that its own dual is the original ordered binary tree.
By definition, if $\alpha$ and $\beta$ are two trees, $(\alpha\vee \beta)^* = \alpha^* \wedge \beta^*$ (for some choice of $\wedge$ and duals $\alpha^*$ and $\beta^*$).
We must show that, conversely, for any choice of $\wedge$ and duals $\alpha^*$ and $\beta^*$, we can choose a dual $(\alpha \wedge \beta)^* = \alpha^* \vee \beta^*$.
We proceed by induction; the result is clear for trees with one leaf.
For the given choice of $\alpha \wedge \beta$, let $\alpha = \gamma \vee \delta$ and $\beta = \mu \vee \nu$ so that $\alpha \wedge \beta = (\gamma \wedge \mu) \vee (\delta \wedge \nu)$.
Then
\begin{eqnarray*}
  (\alpha \wedge \beta)^{*?} &=& \left(\strut (\gamma \wedge \mu) \vee (\delta \wedge \nu) \right)^{*?} \\
  &=& (\gamma \wedge \mu)^{*?} \mathbin{\mathop{\wedge}\limits^?} (\delta \wedge \nu)^{*?} \\
  &=& (\gamma^* \vee \mu^*) \mathbin{\mathop{\wedge}\limits^?} (\delta^* \vee \nu^*) \\
  &=& (\gamma^* \mathbin{\mathop{\wedge}\limits^?} \delta^*) \vee (\mu^* \mathbin{\mathop{\wedge}\limits^?} \nu^*) \\
  &=& (\gamma \vee \delta)^* \vee (\mu \vee \nu)^* \\
  &=& \alpha^* \vee \beta^* ,
\end{eqnarray*}
where $()^{*?}$ indicates that we have not yet chosen the dual, and $\mathop{\wedge}\limits^?$ indicates that we have not yet chosen the union.
The step in which we use $(\gamma \wedge \mu)^* = \gamma^* \vee \mu^*$ and $(\delta \wedge \nu)^* = \delta^* \vee \nu^*$ is the induction step.
In the step where $(\gamma\vee \delta)^* = \gamma^* \wedge \delta^*$ and $(\mu \vee \nu)^* = \mu^* \wedge \nu^*$, we are choosing the $\wedge$ used in the choice of the duals $\alpha^*$ and $\beta^*$.

\section{Application to Geometric Optimization Problems}

One of the main applications of the Fast Tree Transform is the minimization of local first-order geometric energies on an arbitrary graph (of which the harmonic energy is a simple linear example).
The gradient flow of such an energy is a second-order geometric motion~\cite{bib.burchard.tvgeom} (the Laplacian in the case of the harmonic energy).
The local geometric nature of the energy means that the Hessian of the energy has eigenvalues highly correlated with the frequency spectrum of the Laplacian.
Therefore, the FTT can provide a good preconditioner for such optimization problems.

As a preconditioner for conjugate gradients, the FTT by itself results in a minimizer that runs in $\frac32$-power time.
The reason is that the degree of approximate diagonalization provided by the transform is that preconditioning with a diagonal operator in the transform space reduces the effective condition number of the Laplacian to a square root of itself.
The worst case occurs at the low frequency end of the spectrum for a one-dimensional geometry,
where the condition number is reduced from quadratic to linear in the size of the geometry.

To compute the FTT preconditioner, we use a {\em gradient smoothing\/} algorithm (see Section~\ref{sec.gradsmooth} for details).
By using the FTT to apply a low pass filter to the gradient at an increasing logarithmic sequence of frequency cutoffs, we can take accelerated smoothed gradient steps for lower frequency parts of the problem, analogously to a multilevel method.  This compound step, consisting of a logarithmic number of sub-steps, then serves as the preconditioned step for conjugate gradients.

To get a nearly linear time solution, though, we need to combine the FTT with a spanning tree preconditioner.
The reason is that the worst case for the FTT is a one-dimensional geometry, which is however trivial to solve by using the divergence form of the geometric motion to reformulate the problem as a ``current flow'' problem along the one dimension.
More generally, the same ``current flow'' approach allows exact solution in linear time for any tree geometry.
Applying this to a spanning tree in the graph produces a preconditioner that complements the FTT preconditioner.

Existing nearly linear time solutions of geometric optimization problems based on a multilevel approach are quite complex~\cite{bib.spielman.linsolve}.
Recently, a simpler random cycle approach has been demonstrated~\cite{bib.kelner.randcycle}.
All of these have only been applied in the linear case, although it is likely they could be extended to more general geometric motions.

\section{Gradient Smoothing}
\label{sec.gradsmooth}

When using the Fast Tree Transform to precondition gradient descent for a geometric energy, it may not be easy to explicitly compute the correct spectral multiplier for each frequency.
Instead, it would be simpler if we could somehow separately perform the gradient descent for each frequency, or block of similar frequencies.
Then the correct spectral multiplier would effectively come out implicitly, in the length of each partial gradient descent step that we are able to take,
naturally resulting in longer steps for lower frequency parts of the problem and shorter steps for higher frequency parts.

This is the concept behind the {\em gradient smoothing\/} method.
To apply the method, select a logarithmic number of multiplicatively increasing frequency cutoffs that exhaust the spectrum.
Then, for each cutoff in the sequence, smooth the gradient by applying a low pass filter with that cutoff by means of a chosen frequency transform, and perform energy descent in that smoothed direction.
The lower the frequency cutoff, the longer the smoothed gradient step will be, thereby allowing energy in low frequencies to be dissipated as quickly as that in high frequencies.

When FFT is available and used as the frequency transform, the gradient smoothing method is as effective as the multilevel method, but without the complication of constructing a sequence of coarser problems.
The only downside is that there is an extra log factor in the time cost of the solution, because each level of smoothing operates on the full dimensionality of the problem, instead of on an exponentially smaller coarse problem.
When the FFT is not available and we must use the FTT, a gradient smoothing step (consisting of a logarithmic number of sub-steps) can still be used as a preconditioner.

Note that for gradient smoothing to work, the gradient must be represented in the correct geometric space~\cite{bib.burchard.tvgeom}.

\bibliographystyle{amsplain}
\bibliography{document}

\end{document}